\pdfoutput=1

\documentclass[11pt]{article}

\usepackage[final]{acl}

\usepackage{times}
\usepackage{latexsym}

\usepackage[T1]{fontenc}

\usepackage[utf8]{inputenc}

\usepackage{microtype}

\usepackage{inconsolata}

\usepackage{graphicx}
\usepackage{amssymb}
\usepackage{xspace}
\usepackage{xurl}
\usepackage{comment}
\usepackage{pifont}
\usepackage{xcolor}
\usepackage{adjustbox}
\usepackage{booktabs}
\usepackage{lipsum}
\usepackage{tabularx}
\usepackage{makecell}
\usepackage[most]{tcolorbox}
\usepackage{colortbl}
\newtcolorbox{prompt}[2][]{
    colback=white,
    colframe=gray!45,
    fonttitle=\bfseries,
    coltitle=black,
    title=#2,
    #1,breakable
}
\usepackage{fontawesome5}

\usepackage{cleveref}
\crefname{section}{§}{§§}

\newcommand{\tabitem}{~~\llap{\textbullet}~~}

\newcommand{\eg}{e.g.,\xspace}
\newcommand{\ie}{i.e.,\xspace}

\newcommand{\modelName}{\textsc{Thanos}\xspace}
\newcommand{\modelEmoji}{\includegraphics[height=.9em,trim=0 .4em 0 0]{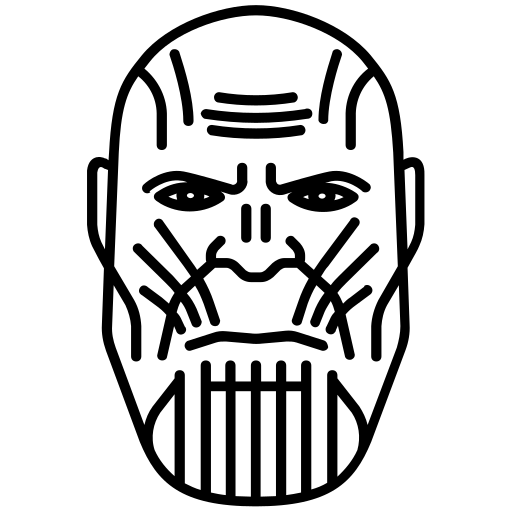}}
\newcommand{\model}{\modelEmoji \xspace \modelName \xspace}

\newcommand{\datasetName}{\textsc{Multifaceted Skill-of-Mind}\xspace}

\newcommand{\socialcontextEmoji}{\includegraphics[height=1.1em,trim=0 .4em 0 0]{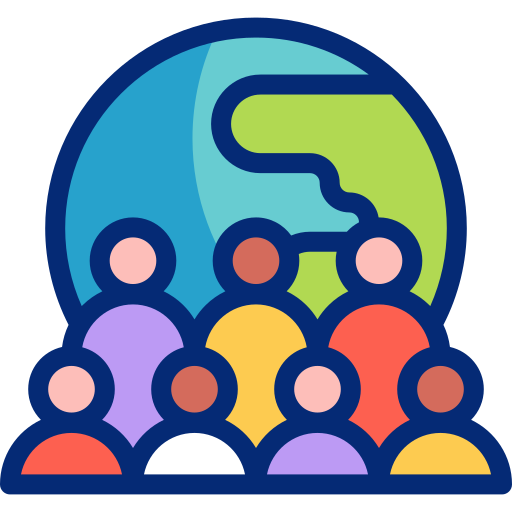}}
\newcommand{\dialogueEmoji}{\includegraphics[height=1.1em,trim=0 .4em 0 0]{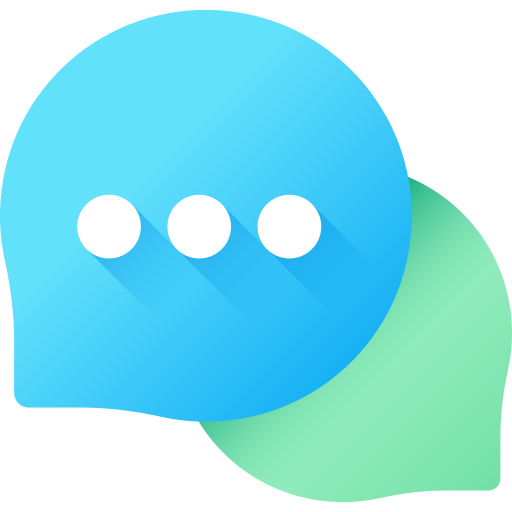}}
\newcommand{\skillEmoji}{\includegraphics[height=1.1em,trim=0 .4em 0 0]{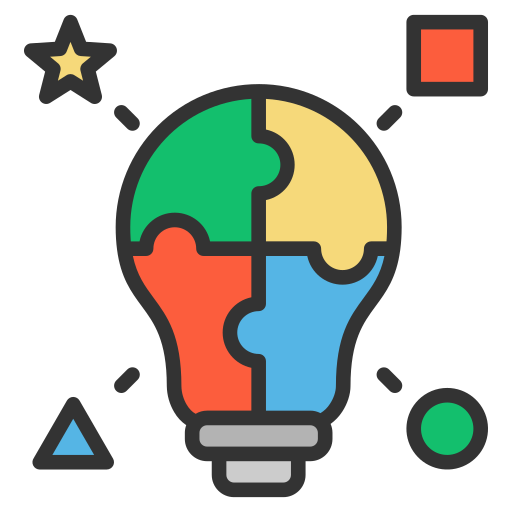}}

\newcommand{\purplerow}{\rowcolor[HTML]{dcbbfe}}

\newcommand{\system}{System Message}
\newcommand{\user}{Instruction}

\newcommand{\flask}{FLASK}
\newcommand{\bst}{BlendedSkillTalk}
\newcommand{\bsbt}{Blended Skill BotsTalk}
\newcommand{\bsbtAbbv}{BS$\mathbb{B}$T}
\newcommand{\stark}{\textsc{Stark}}
\newcommand{\soda}{\textsc{Soda}}
\newcommand{\prosocial}{\textsc{ProsocialDialogue}}
\newcommand{\empathy}{\textsc{EmpatheticDialogues}}
\newcommand{\convchron}{\textsc{ConversationChronicles}}
\newcommand{\wizard}{Wizard of Wikipedia}
\newcommand{\cactus}{\textsc{Cactus}}
\newcommand{\casino}{\textsc{CaSiNo}}
\newcommand{\multiwoz}{MultiWOZ 2.2}
\newcommand{\persuasion}{\textsc{PersuasionForGood}}
\newcommand{\pearl}{\textsc{Pearl}}
\newcommand{\persona}{\textsc{SynPersonaChat}}
\newcommand{\daily}{\textsc{DailyDialog}}

\newcommand{\gemma}{\texttt{Gemma-2-2B}}
\newcommand{\llama}{\texttt{LLaMA-3.1-8B}}
\newcommand{\cosmo}{\texttt{COSMO-XL}}

\newcommand{\github}[1]{%
   \href{#1}{\faGithub}%
}

\title{\model: Enhancing Conversational Agents with \textit{Skill-of-Mind}-Infused Large Language Model}

\author{
Young-Jun Lee \textsuperscript{\rm 1} \quad
Dokyong Lee \textsuperscript{\rm 2} \quad
Junyoung Youn \textsuperscript{\rm 2} \quad
Kyeongjin Oh \textsuperscript{\rm 2} \quad 
\textbf{Ho-Jin Choi} \textsuperscript{\rm 1} 
\\
\textsuperscript{\rm 1} School of Computing, KAIST \quad
\textsuperscript{\rm 2} KT Corporation
\\
\texttt{\{yj2961, hojinc\}@kaist.ac.kr} \quad
\texttt{\{dokyong.lee, junyoung.youn, kyeong-jin.oh\}@kt.com}
}

\begin{document}
\maketitle

\begin{abstract}
    To increase social bonding with interlocutors, humans naturally acquire the ability to respond appropriately in a given situation by considering which conversational skill is most suitable for the response — a process we call \textit{skill-of-mind}~\footnote{We derive the term \textit{skill-of-mind} from \textit{theory of mind}~\cite{premack1978does}, which refers to the ability to understand and infer others' mental states, intentions, and beliefs from situational descriptions (\eg narratives). Skill-of-mind refers to interpreting/understanding the current conversational context based on social dynamics (\eg demographics, persona) and optimizing social interaction through conversational skills.}. For large language model (LLM)-based conversational agents, planning appropriate conversational skills, as humans do, is challenging due to the complexity of social dialogue, especially in interactive scenarios. To address this, we propose a \textit{skill-of-mind}-annotated conversation dataset, named \datasetName, which includes multi-turn and multifaceted conversational skills across various interactive scenarios (\eg long-term, counseling, task-oriented), grounded in diverse social contexts (\eg demographics, persona, rules of thumb). This dataset consists of roughly 100K conversations. Using this dataset, we introduce a new family of \textit{skill-of-mind}-infused LLMs, named \model, with model sizes of 1B, 3B, and 8B parameters. With extensive experiments, these models successfully demonstrate the \textit{skill-of-mind} process and exhibit strong generalizability in inferring multifaceted skills across a variety of domains. Moreover, we show that \model significantly enhances the quality of responses generated by LLM-based conversational agents and promotes prosocial behavior in human evaluations. \github{} Code: \url{https://github.com/passing2961/Thanos}.
\end{abstract}

\section{Introduction} \label{main_sec:intro}

\begin{figure}[t]
    \centering
    \includegraphics[width=0.9\linewidth]{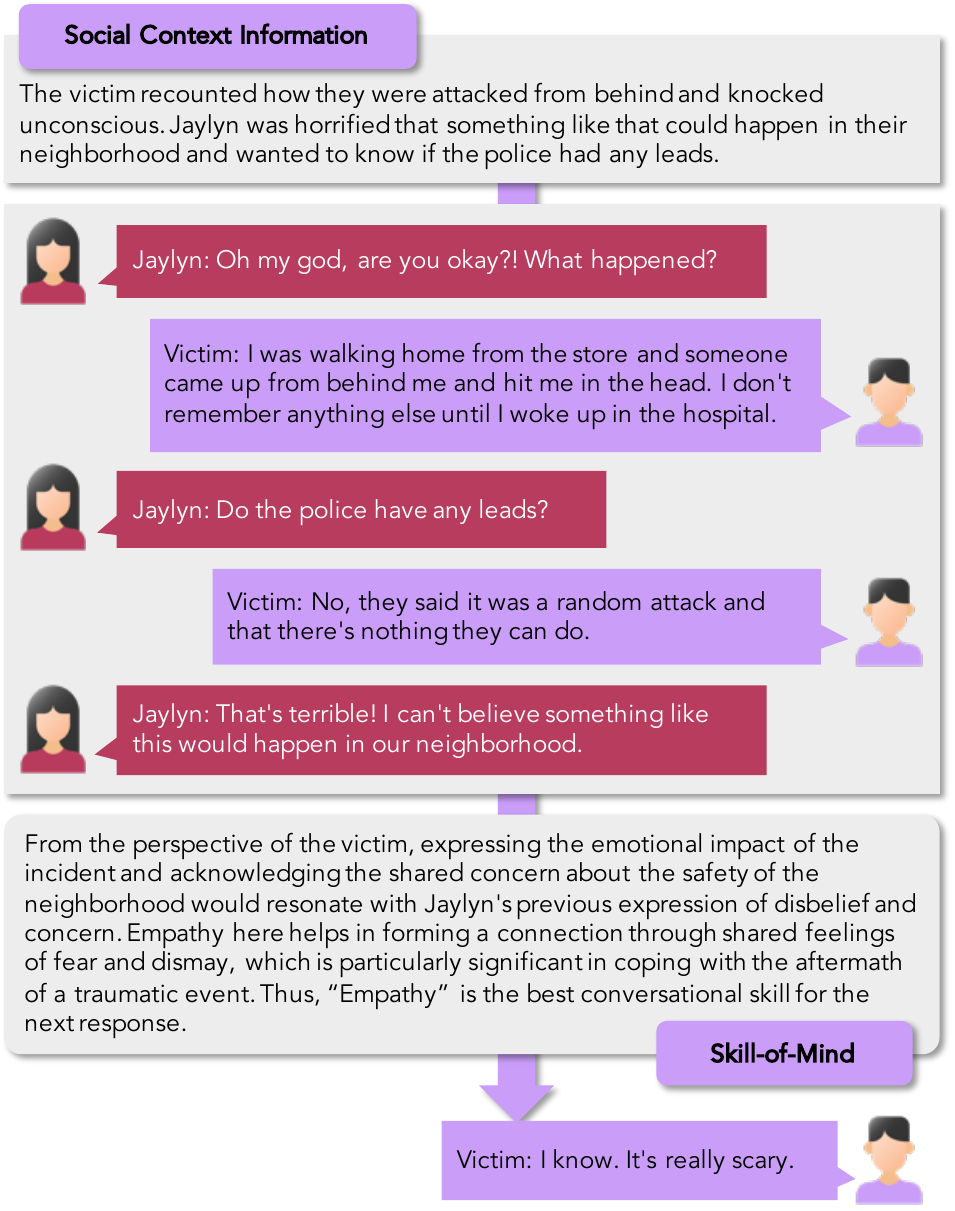}
    \caption{An overview of \textit{skill-of-mind} process.}
    \label{main_fig:teaser}
\end{figure}

In everyday conversations, humans engage in diverse and complex interactions with their interlocutors (\eg friends, colleagues) by understanding and interpreting their interlocutors' situations~\cite{rashkin2018towards,lee2022does} and personas~\cite{zhang2018personalizing,lee2022personachatgen}, and by recalling memorable events or moments~\cite{bae2022keep,jang2023conversation,lee2024stark}. Humans do not always know the most appropriate response for every turn in a multi-turn conversation. Instead, they learn to choose suitable responses by internally considering the most appropriate conversational skills at that moment, based on the interlocutor and social context over time. As illustrated in Figure~\ref{main_fig:teaser}, for example, humans reflect on which skill to use for the next turn by internally reasoning about which skill would be most appropriate. This process evolves through self-reflection and feedback, as people assess the positive or negative reactions of their interlocutors. We refer to this entire process as \textit{skill-of-mind}, which involves interpreting and understanding the current dialogue situation, planning the best skill strategy for the next response, and then selecting the most appropriate conversational skill.

Recently, conversational agents powered by LLMs~\cite{touvron2023llama,llama3modelcard,team2024gemma} have demonstrated impressive capabilities in logical reasoning~\cite{pan2023logic} and creativity~\cite{franceschelli2023creativity}. However, these agents still struggle with social commonsense reasoning~\cite{chae2023dialogue} and strategic communication skills in an interactive environment~\cite{zhou2023sotopia}. Additionally, despite interacting with hundreds of millions of individual users~\cite{zhao2024wildchat}, each possessing a distinct perspective or persona, current agents do not effectively personalize their responses to users~\cite{lee2024aligning}. We posit that directly generating the next response is particularly challenging for LLM-based conversational agents due to the complexity of social dialogue, specifically: (1) the \textit{one-to-many} problem~\cite{li2015diversity,bao2019plato}, where multiple plausible responses exist, and (2) the scattering of key evidence~\cite{chae2023dialogue}, where critical information or phrases are dispersed throughout the conversation. To address this, we suggest that before generating the next response, interpreting the dialogue situation and planning the most plausible conversational skill — similar to how humans' internal mind states operate — can enhance response quality, even in interactive situations involving social dynamics, by acting as a form of guidance.

To this end, we introduce \datasetName, a collection of multi-turn, multifaceted \textit{skill-of-mind} annotated conversations. This dataset includes annotations for both explanations and conversational skills, covering a wide range of conversational skills across one-sided turns within dialogues. The dataset is derived from 12 existing source dialogue datasets, which encompass diverse social contexts and scenarios (\eg chitchat, counseling). To annotate \textit{skill-of-mind}, we prompt GPT-4~\cite{achiam2023gpt} (\ie \texttt{gpt-4-turbo}) to generate explanations and identify conversational skills from a predefined collection of conversational skills, organized hierarchically into five main categories: Interpersonal, Memory \& Knowledge Management, Cognitive \& Problem-Solving, Communication \& Listening, Task-Oriented.

Additionally, we propose a new family of \textit{skill-of-mind}-infused LLMs, \model, which generate both an explanation and the most appropriate conversational skill in LLM-based conversational agents. Through extensive experiments, \model accurately predicts conversational skills and generates high-quality explanations across various dialogue scenarios, demonstrating strong \textit{generalizability} in skill prediction. To further validate the effectiveness of \model, we incorporate the generated \textit{skill-of-mind} as augmented input prompts for LLM-based conversational agents when responding to the next turn, resulting in significant improvements in response quality.

Our contributions are summarized as follows: (1) We introduce a new social concept, \textit{skill-of-mind}, which involves interpreting dialogue situations, planning the best skill strategy, and selecting the appropriate conversational skill. (2) We present a multi-turn, multifaceted \textit{skill-of-mind}-annotated conversation dataset, \datasetName, which encompasses diverse social dynamics and interactive scenarios. (3) Using \datasetName, we propose a family of \textit{skill-of-mind}-infused LLMs, \model, with model sizes of 1B, 3B, and 8B parameters. With extensive experiments, we demonstrate the effectiveness and generalizability of \model across various scenarios.

\section{\datasetName} \label{main_sec:dataset}

In this section, we explain the definition of \textit{skill-of-mind} (\cref{sec:definition}), why predicting \textit{skill-of-mind} is necessary (\cref{sec:why_skill}), the hierarchical taxonomy of conversational skills (\cref{sec:ingredient}), and the data construction process (\cref{sec:dataset_construction}). Lastly, we analyze the constructed dataset (\cref{sec:analysis}).

\subsection{Definition of ``Skill-of-Mind''} \label{sec:definition}
We begin by defining what ``skill-of-mind'' means in the context of this work. In the conversational AI literature~\cite{smith2020can,kim2022botstalk}, these ``skills'' are sometimes regarded as communication strategies~\cite{zhou2023sotopia}, but generally refer to the desirable abilities required to maintain continuous and meaningful conversations with an interlocutor. These skills cover a broad spectrum, ranging from general abilities (\eg empathy, persona) to task-specific functions (\eg making phone calls, booking a hotel). In everyday social interactions, which are a core component of human life, people typically reflect on the current situation and consider who the interlocutor is before selecting the most appropriate skill. This internal cognitive process is often represented as an explanation/rationale~\cite{zhou2021think,wei2022chain}. These explanations and skill choices are significantly influenced by the characteristics of the interlocutor, such as demographics, persona, and relationship. 

\subsection{Why ``Skill-of-Mind'' is necessary?} \label{sec:why_skill}

At the heart of the conversation is social interaction~\cite{myllyniemi1986conversation}, a domain where current LLMs have limited understanding and struggle to effectively handle social interactive scenarios~\cite{zhou2023sotopia,liu2023training}. As a result, generating more engaging and natural responses directly through LLM-based conversational agents is challenging. This is because LLMs are primarily designed to solve complex reasoning tasks as general agents through alignment tuning~\cite{ouyang2022training,chung2024scaling}, making them ill-suited to function as social dialogue agents. By introducing guidance based on the concept of \textit{skill-of-mind}, LLM-based conversational agents can more effectively navigate social interactions. Current LLMs demonstrate better alignment and are capable of following user queries, so grounding responses in \textit{skill-of-mind} can help narrow down the response options and focus on skill-specific aspects. This leads to more accurate outputs by reducing the range of possible responses (\ie one-to-many problem). In practical applications, most conversational systems adopt a modular approach~\cite{lee2023prompted} where multiple skill-specialized dialogue agents are used to strengthen user interaction. In these systems, several skill-specific agents generate responses simultaneously, and a re-ranking module~\cite{bae2022building} selects the best one for the given dialogue situation. While this approach is effective, it is also resource-intensive. In real-world scenarios, where multiple conversational skills are required, even with parallel processing, inference may not be as fast as using a single skill-expert agent. By focusing on a single agent grounded in \textit{skill-of-mind}, we can potentially improve inference speed and reduce latency.

\subsection{Ingredients for ``Skill-of-Mind''} \label{sec:ingredient}

The concept of ``skill-of-mind'' consists of three main components: (1) social context, (2) explanation, and (3) conversational skill, which are described as follows.

\paragraph{Social Context Information.} Socially interactive dialogues involve a wide range of social dynamics, such as demographics, personal experiences, and relationships. We believe that these factors influence the \textit{skill-of-mind}. For instance, emotional empathy is more appropriate in conversations with a significant romantic partner than with an AI teaching mentor. Therefore, when considering social context information, we take into account various elements, such as the current situation or narrative, social relationships, persona, memory, and past dialogue histories.~\footnote{Note that we do not generate this information from scratch; rather, it originates from the source dialogue dataset used in this work.}

\paragraph{Explanation/Rationale.} This involves interpreting and understanding the current situation to determine the most optimized conversational skill for generating an engaging response that strengthens social rapport~\cite{zech2005talking} with the interlocutor in the given dialogue. To achieve a higher quality of explanation, similar to how a human would respond, we adopt \textit{perspective-taking}-style~\cite{davis1983measuring,ruby2004would,kim2021perspective}, which prompts GPT-4 to imagine the actual speaker in the dialogue. The explanation is represented in free-form sentences.

\paragraph{Conversational Skill.} As discussed in \cref{sec:definition}, there is a broad range of conversational skills in real-world scenarios, such as empathy or persona management in chitchat, hotel reservations in task-oriented dialogues, and memory recall in long-term conversations. To account for this spectrum, we design a taxonomy that covers \textit{fine-grained} levels of conversational skills across diverse scenarios. We develop a hierarchical taxonomy that encompasses distinct and \textit{non-overlapping}~\footnote{In reality, it is not always necessary to assign each conversational skill to a single category since many skills share overlapping characteristics. For instance, the skill of ``Active Listening'' can also be categorized under ``Interpersonal Skills.'' However, for clarity in this work, we structure the taxonomy with \textit{non-overlapping} conversational skill categories.} categories of skills. At the first level, we identify five main categories: (1) Interpersonal Skills, (2) Memory \& Knowledge Management Skills, (3) Cognitive \& Problem-Solving Skills, (4) Communication \& Listening Skills, and (5) Task-Oriented Skills.

\begin{itemize}
    \item \textbf{Interpersonal Skills:} These skills are essential for enhancing social interaction by requiring a deep understanding of the interlocutor's emotional state and adapting to their personality or relationship dynamics for more seamless and engaging communication. They also involve demonstrating prosocial behavior in problematic situations. We also consider image-sharing behavior, which frequently occurs via instant messaging tools. This category includes Empathy, Personal Background, Persona Recall, Self-Disclosure, Negotiation, Conflict Resolution, Conflict Avoidance, Persuasion, Commonsense Understanding, Cultural Sensitivity, Ethics, Harmlessness, Avoiding Social Bias, Helpfulness, Mentoring, Image Commenting, and Image Sharing.

    \item \textbf{Memory \& Knowledge Management Skills:} These skills are primarily used to provide knowledgeable responses by sharing or acquiring information and recalling memories, which is important for maintaining long-term communication, particularly in senior care services~\cite{bae2022keep,bae2022building}. This category includes Memory Recall, Knowledge Sharing, Knowledge Acquisition, and Knowledge Searching.

    \item \textbf{Cognitive \& Problem-Solving Skills:} These skills are required for solving complex problems or performing factual reasoning tasks. This category includes Critical Thinking, Logical Thinking, Creative Problem Solving, Factual Problem Solving, and Decision-Making.
    
    \item \textbf{Communication \& Listening Skills:} Effective listening is critical in the communication process~\cite{main1985sell,castleberry1993effective}. Therefore, we include these skills in our taxonomy, which encompasses Clarification, Confirmation, Rephrasing, Echoing, Topic Transition, Rhetoric, Active Listening, Reflective Listening, and Immediate Response.
    
    \item \textbf{Task-Oriented Skills:} In practical scenarios, humans often request conversational agents (\eg Alexa~\footnote{\url{https://developer.amazon.com/en-US/alexa}}) to perform tasks such as hotel or restaurant reservations, provide weather information, or offer movie recommendations. We also consider these skills, which include Recommendation, Task Execution, and Urgency Recognition.
    
\end{itemize}

\subsection{Dataset Construction Process} \label{sec:dataset_construction}

Based on the above ingredients of \textit{skill-of-mind} (\cref{sec:ingredient}), we first collect source dialogue datasets and annotate them with \textit{skill-of-mind}, resulting in \datasetName.

\paragraph{Step 1: Source Dataset Collection.} To build more flexible and versatile \textit{skill-of-mind}-infused LLM, as a source data, we collect 12 multi-turn dialogue datasets, which are publicly available online: \soda~\cite{kim2022soda}, \convchron~\cite{jang2023conversation}, \prosocial~\cite{kim2022prosocialdialog}, \empathy~\cite{rashkin2018towards}, \wizard~\cite{dinan2018wizard}, \cactus~\cite{lee2024cactus}, \casino~\cite{chawla2021casino}, \multiwoz~\cite{zang2020multiwoz}, \persuasion~\cite{wang2019persuasion}, \pearl~\cite{kim2024pearl}, \persona~\cite{jandaghi2023faithful}, and \stark~\cite{lee2024stark}. In total, we collect source dialogues from the training sets. We then split each dialogue into sub-dialogues by focusing on one-sided exchanges. For example, given a dialogue $\mathcal{D} = \{(s_i, u_i)\}_{i=1}^4$, we create two sub-dialogues: $\mathcal{D}_1 = \{(s_i, u_i)\}_{i=1}^2$ and $\mathcal{D}_2 = \{(s_i, u_i)\}_{i=1}^4$. We remove sub-dialogues with fewer than four turns, as we believe that early in the dialogue, there is a higher distribution of non-informative skills, such as greetings, rather than informative skills. We then randomly sample sub-dialogues from each source dataset in specific proportions. As a result, we obtain a total of 100K dialogues.

\paragraph{Step 2: Annotating \textit{Skill-of-Mind}.} We prompt GPT-4~\cite{achiam2023gpt} (\ie \texttt{gpt-4-turbo}) to annotate \textit{skill-of-mind} into the collected source dialogues. Specifically, it provides internal reasoning about which skills are appropriate for the next turn response in the dialogue and identifies the relevant multifaceted conversational skills from the predefined taxonomy (\cref{sec:ingredient}), taking into account the interlocutor’s perspective (\ie perspective-taking). Each instance in \datasetName consists of three input components (social context information, dialogue, next response) and two output components (explanation, skill). 

The input components are described as follows:
\begin{itemize} 
    \item \textbf{Dialogue:} A dialogue between two speakers from the collected source datasets in step (1). 
    \item \textbf{Next Response:} The next response in the dialogue, which should align with the relevant explanation and conversational skill. Given the subjective nature of dialogue, if only the dialogue is provided without a golden response, GPT-4 can still generate plausible explanations and skills that are not compatible with the natural flow of the original dialogue. 
    \item \textbf{Social Context Information:} Social context information encompasses various social dynamics, which vary depending on the source dialogues. For example, this includes social narratives in \soda \xspace and demographic factors (\eg age, gender, birthplace, residence), personal narratives, or past session dialogue summaries in \stark.
\end{itemize}

The output components are described as follows:
\begin{itemize} 
    \item \textbf{Explanation:} A rationale explaining which skill is necessary to maintain continuous interaction with the interlocutor, given the input dialogue and the next response. To create more realistic explanations, GPT-4 is induced to engage in a perspective-taking process. 
    
    \item \textbf{Conversational Skill:} Based on the explanation, one or more conversational skills relevant to the next response are selected from the predefined skill collections.
\end{itemize}

The prompt template is presented in Appendix~\ref{supp_sec:prompt}.We instruct GPT-4 to produce a structured output in \texttt{JSON} format, excluding any cases that fail to parse correctly. In total, we obtain 99,997 annotations (approximately 100K) and split the data into a 9:1 train/test ratio for evaluation. In instances with multiple \textit{skill-of-mind} annotations, we randomly select one for training \model. An example of \textit{skill-of-mind} annotation is presented in Table~\ref{main_tab:dataset_example}.

\subsection{Analysis} \label{sec:analysis}

{\renewcommand{\arraystretch}{1.35}
\begin{table}[!t]
\centering
\begin{adjustbox}{width=\linewidth}
\begin{tabular}{@{}lcccc@{}}
\toprule
Dataset & Train? & Explanation? & \# of D. & \# of S. \\ \midrule
BST~\cite{smith2020can}     &   \ding{51}     &     \ding{55}         &    6,808            &   3          \\
\bsbtAbbv~\cite{kim2022botstalk}   &   \ding{51}     &      \ding{55}        &  300,000              &      3       \\
\flask~\cite{ye2023flask} &   \ding{55}     &      \ding{55}        &      1,740          &      12       \\
\purplerow \datasetName    &   \ding{51}     &     \ding{51}         &      99,997          &  38+           \\ \bottomrule
\end{tabular}
\end{adjustbox}
\caption{Comparison of \datasetName with existing datasets regarding skills: \bst \xspace (BST), \bsbt \xspace (\bsbtAbbv), and \flask. D. and S. denote the dialogue and skill, respectively. The ``+'' next to 38 indicates the potential for additional skills as GPT-4 sometimes generates new skills not present in the predefined collection (\eg \textit{feedback giving}). To build a more flexible model, we include these cases in our training dataset.}
\label{main_tab:dataset_comparison}
\vspace{-1em}
\end{table}}

\paragraph{Comparison to Existing Datasets.} In Table~\ref{main_tab:dataset_comparison}, we compare \datasetName with other existing datasets that include certain skills. In summary, \datasetName is the first dataset to contain both explanations and skills. Although the number of dialogues is smaller than that of the \bsbtAbbv\xspace dataset, we include a greater number of conversational skills, which enhances the generalizability of the trained model. Compared to \flask, our dataset also includes a larger variety of skills, whereas \flask is designed to evaluate fine-grained LLM capabilities and primarily focuses on instruction-based skills. In contrast, our dataset offers a comparable dialogue size, and a substantial number of skills, and includes both explanations and skills, making it a robust resource for generalizable skill prediction.

\begin{figure}[t]
    \centering
    \includegraphics[width=0.9\linewidth]{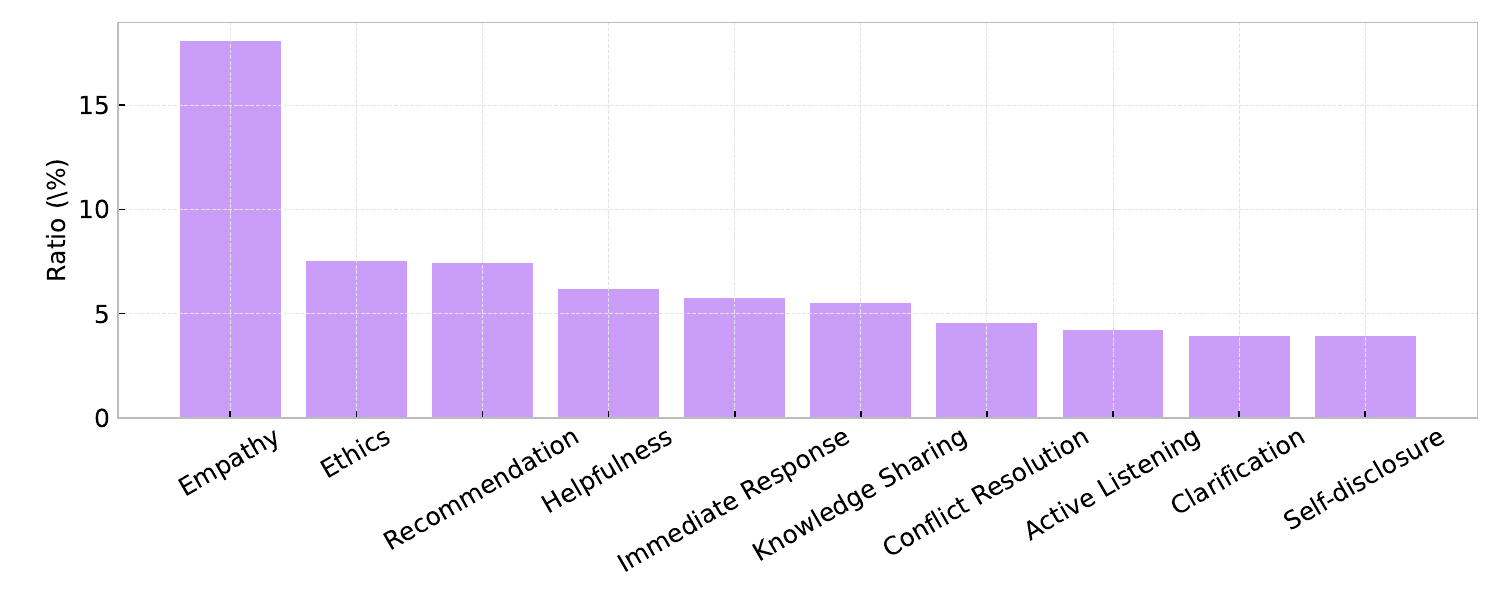}
    \caption{The ration (\%) of Top-10 conversational skill categories in \datasetName.}
    \label{main_fig:skill_dist}
\end{figure}

\paragraph{Distribution of \textit{Skill-of-Mind}.} 

Figure~\ref{main_fig:skill_dist} shows the distribution of the Top-10 conversational skill categories generated by GPT-4 (as discussed in~\cref{sec:dataset_construction}). We analyzed a total of 109,591 \textit{skill-of-mind} annotations (considering multiple annotations per dialogue). The most prominent skill is \texttt{Empathy}, likely due to the presence of socially interactive datasets, such as \soda, \convchron, and \empathy, which contain a large proportion of empathetic interactions — crucial in social dialogues. Additionally, \texttt{Ethics} and \texttt{Helpfulness} also occupy significant portions, indicating that the model trained on our dataset may serve as a \textit{safeguard} to promote prosocial behavior.

\paragraph{Human Evaluation.}

To assess the quality of \datasetName, we conduct a human evaluation based on five criteria: (1) relevance, (2) plausibility, (3) understanding, (4) skill alignment, and (5) skill adequacy. Each human evaluator rates 100 randomly sampled instances on a 4-point Likert scale for all criteria. Criteria (1-3) measure the quality of the \textit{explanation}, while criteria (4-5) assess the quality of the \textit{conversational skill}. Further details about each evaluation criterion and the recruitment process for human evaluators are provided in the Appendix~\ref{supp_sec:human_eval_question} and Appendix~\ref{supp_sec:detail_human_eval}. On average, we achieve notably high scores: 3.72 for relevance, 3.75 for plausibility, 3.74 for understanding, 3.64 for skill alignment, and 3.59 for skill adequacy. Additionally, we compute inter-rater agreement (IA) using Krippendorff's $\alpha$, yielding a value of 0.62, which indicates a substantial level of agreement. These results demonstrate the reliability and quality of \datasetName, particularly with respect to generating human-like \textit{skill-of-mind} in interactions.

{\renewcommand{\arraystretch}{1}
    \begin{table}[t] \begin{center}
    \small
    \setlength{\tabcolsep}{3pt}
    \begin{tabularx}{\linewidth}{X}
        \toprule
        \socialcontextEmoji \xspace\xspace \textbf{Social Context Information}: \\
        Sawyer tells Gregg that he needs help with groceries tonight since he has to cook dinner for everyone. They go together and purchase the items on the list including chicken, broccoli, and pasta. \\
        \midrule
        \dialogueEmoji \xspace\xspace \textbf{Dialogue}:\\
        \textbf{Sawyer:} Hey Gregg, do you think you could help me out with groceries tonight? I've got to cook dinner for everyone and it's a bit of a hassle by myself. \\
        \textbf{Gregg:} Sure thing, Sawyer. What do you need to get? \\
        \textbf{Sawyer:} Well, let's see...we need chicken, broccoli, and pasta. Oh, and some bread too. \\
        \textbf{Gregg:} Alright, sounds like we can make a pretty good meal out of that. Let's head to the store then. \\
        \textbf{Sawyer:} Thanks, Gregg. I really appreciate it. \\
        \textbf{Gregg:} So, Sawyer, how come you're cooking dinner tonight? \\
        \textbf{Sawyer:} Oh, you know...just felt like it. Plus, I love to cook so it's not a big deal. \\
        \textbf{Gregg:} That's really nice of you. I'm sure everyone will appreciate it. \\
        \textbf{Sawyer:} Yeah, I hope so! I just want to make something that everyone will enjoy. \\
        \midrule
        \skillEmoji \xspace\xspace \textbf{Skill-of-Mind}:\\ 
        \tabitem \textbf{Explanation}: In responding to Sawyer, I want to show that I've been attentively listening to his thoughts and appreciating the effort he is putting into cooking dinner. By affirming his efforts and expressing confidence that everyone will enjoy his cooking, I am validating his feelings and intentions, which is key in making him feel supported. \\
        \tabitem \textbf{Conversational Skill}: Active Listening \\
        \bottomrule
    \end{tabularx}
    \vspace{-5pt}
    \caption{
        A sample from \datasetName. 
    }
    \vspace{-15pt}
    \label{main_tab:dataset_example}
\end{center}\end{table}}

\section{\model: Skill-of-Mind-infused LLM} \label{main_sec:thanos}

\paragraph{Backbone LLM.} To enhance performance across various applications, we introduce a new family of skill-of-mind-infused LLMs with varying model sizes: \model-$\{1, 3, 8\}$B. For \modelName 1B, we fine-tune LLaMA-3.2-1B-Instruct; for \modelName 3B, we fine-tune LLaMA-3.2-3B-Instruct; and for \modelName 8B, we fine-tune LLaMA-3.1-8B-Instruct using \datasetName. 

\paragraph{Input \& Output.} During \modelName training, we provide social context information and dialogue as input prompts. Since each source dialogue in our dataset contains varying levels of social context information, we design source-specific social context prompt templates. For each source dialogue, we create five different social context prompt templates, randomly sampling one during training for flexible generation. The social context prompt templates are presented in the Appendix~\ref{supp_sec:social_context}. For the output, \modelName is trained to sequentially generate an explanation, followed by conversational skill, similar to a Chain-of-Thought~\cite{wei2022chain} fine-tuning approach. To mitigate degeneration issues, we introduce a \texttt{[RESULT SKILL]} token between the explanation and the conversational skill, as seen in prior work~\cite{kim2023prometheus}.

\paragraph{Implementation Details.} We fine-tune \modelName-$\{1, 3, 8\}$B using LoRA~\cite{hu2021lora} and PyTorch Fully-Sharded Data Parallel (FSDP). LoRA is applied to all linear layers with a rank of 256 and an alpha of 256. We set the maximum number of epochs to 3, with a batch size of 8 per GPU, using a StepLR scheduler and a learning rate of 1e-5 with the AdamW optimizer. All experiments are conducted on 8 NVIDIA A100 GPUs (40 GB).

\section{Experiments} \label{main_sec:expr}

\subsection{Experimental Setup}

\paragraph{Task Definition.} To evaluate the effectiveness of \modelName, we conduct three tasks: (1) \textit{Skill Classification}: assessing the accuracy of skills generated by LLM-based agents and \model across various dialogue scenarios; (2) \textit{Explanation Generation}: evaluating the quality of explanations produced in these scenarios; and (3) \textit{Response Generation}: determining whether the generated skills and explanations improve the response quality in LLM-based agents.

\paragraph{Evaluation Datasets.} We use different evaluation datasets for each of the three tasks: For (1) \textit{Skill Classification}, we use \bst\xspace (BST)~\cite{smith2020can}, \bsbt\xspace (\bsbtAbbv)~\cite{kim2022botstalk}, \datasetName, and PhotoChat~\cite{zang2021photochat} to assess image-sharing behavior in multi-modal interactions, and \prosocial~\cite{kim2022prosocialdialog} for skills related to prosocial behavior, acting as a safeguard; For (2) \textit{Explanation Generation}, we use \datasetName; For (3) \textit{Response Generation}, we use out-of-domain dialogue datasets like \daily~\cite{li2017dailydialog}, following previous methods~\cite{kim2022soda,chae2023dialogue}, and for in-domain settings, we use \textsc{Soda} and \textsc{ProsocialDialogues} to evaluate how well \model promotes prosocial behavior.

{\renewcommand{\arraystretch}{1.35}
\begin{table*}[!t]
\centering
\begin{adjustbox}{width=0.8\linewidth}
\begin{tabular}{@{}lcccccccccc@{}}
\toprule
                 & \multicolumn{5}{c}{Skill Classification}                                                          & \multicolumn{5}{c}{Explanation Generation}                                                   \\ \cmidrule(l){2-11} 
Models       & Ours           & Prosocial      & BST            & PhotoChat      & Avg.           & B-1            & B-2            & B-4           & R-L            & BertScore      \\ \midrule
\gemma & 3.30           & 12.70          & 1.97           & 5.06           & 5.76           & 8.00           & 2.40           & 0.30          & 4.30           & 29.33          \\
\llama  & 5.20           & 18.42          & 7.10           & 4.03           & 8.69           & 10.90          & 3.90           & 0.70          & 3.80           & 26.18          \\
\purplerow \model 1B       & 27.50          & 50.40          & 24.16          & 12.60          & 28.67          & 29.80          & 14.10          & 4.20          & 18.70          & 88.49          \\
\purplerow \model 3B       & 28.80          & 50.80          & \textbf{24.85} & 14.87          & 29.83          & 30.90          & 14.70          & 4.90          & 19.50          & 88.45          \\
\purplerow \model 8B       & \textbf{29.70} & \textbf{53.80} & 23.08          & \textbf{15.81} & \textbf{30.60} & \textbf{31.20} & \textbf{15.10} & \textbf{5.40} & \textbf{20.10} & \textbf{88.53} \\ \bottomrule
\end{tabular}
\end{adjustbox}
\caption{Comparison of skill classification accuracy (\%) and explanation generation on \datasetName (Ours), \prosocial\xspace (Prosocial), BST, and PhotoChat datasets. B-1/2/4 refer to BLEU-1/2/4~\cite{papineni2002bleu}, and R-L refers to ROUGE-L~\cite{lin2004rouge} for simplicity.}
\label{main_tab:skill_perf}

\end{table*}}

\paragraph{Baselines.} Since our goal is to enhance the quality and sophistication of responses generated by open-sourced LLM-based conversational agents through the incorporation of \textit{skill-of-mind} capabilities, we select three different, widely-adopted LLM-based conversational agents as baselines: 1) \gemma, 2) \llama\xspace (which also serves as the backbone for our \modelName 8B), and 3) \cosmo~\cite{kim2022soda} which is specialized to social dialogue. 

\paragraph{Evaluation Metrics.} For (1), we measure accuracy (\%). For (2), we evaluate using BLEU-1/2/4~\cite{papineni2002bleu} and ROUGE-L~\cite{lin2004rouge}. For (3), we apply the same metrics as in (2). However, due to the inherently subjective nature of dialogue evaluation, in addition to BLEU and ROUGE, we conduct a human evaluation based on four criteria: (1) naturalness, (2) engagingness, (3) consistency, and (4) overall quality.

\subsection{Results}

{\renewcommand{\arraystretch}{1.35}
\begin{table*}[!t]
\centering
\begin{adjustbox}{width=\linewidth}
\begin{tabular}{@{}lccccccccccccccc@{}}
\toprule
                 & \multicolumn{5}{c}{\daily}                                                  & \multicolumn{5}{c}{\empathy}                                                      & \multicolumn{5}{c}{\prosocial}                                                    \\ \cmidrule{2-16}
Models      & B-1            & B-2           & B-4           & R-L            & BertScore      & B-1            & B-2           & B-4           & R-L            & BertScore      & B-1            & B-2           & B-4           & R-L            & BertScore      \\ \midrule
\gemma & 6.02           & 2.28          & 0.57          & 11.53          & 82.71          & 5.75           & 1.76          & 0.37          & 10.12          & 84.03          & 13.77          & 4.32          & 0.76          & 12.25          & 85.53          \\
\purplerow + \modelName 1B        & 10.55          & 3.66          & 0.87          & 11.66          & 83.86          & 13.37          & \textbf{4.1}  & \textbf{1.01} & 11.38          & 85.47          & 18.48          & 5.66          & 0.95          & 12.63          & 86.2           \\
\purplerow + \modelName 3B        & 11.01          & 3.63          & 0.82          & 11.74          & 83.72          & 13.68          & 4.03          & 0.87          & 11.56          & 85.52          & 18.51          & 5.8           & \textbf{1.07} & \textbf{12.98} & 86.25          \\ 
\purplerow + \modelName 8B        & \textbf{11.16} & \textbf{3.94} & \textbf{0.97} & \textbf{12.46} & \textbf{83.82} & \textbf{13.67} & 4.01          & 0.76          & \textbf{11.76} & \textbf{85.54} & \textbf{18.54} & \textbf{5.84} & 1.01          & 12.8           & \textbf{86.26} \\ \midrule
\llama  & \textbf{11.6}  & \textbf{4.44} & \textbf{1.16} & \textbf{11.5}  & 83.51          & 9.74           & 3.02          & 0.6           & 10.42          & 84.73          & 15.12          & 5.49          & 1.24          & 13.07          & 85.93          \\
\purplerow + \modelName 1B        & 9.93           & 3.49          & 0.85          & 10.82          & 83.38          & 10.63          & \textbf{3.18} & 0.55          & 10.65          & 84.88          & \textbf{17.83} & \textbf{6.03} & \textbf{1.27} & 13.17          & 86.01          \\
\purplerow + \modelName 3B        & 9.95           & 3.42          & 0.79          & 10.97          & 83.42          & 10.94          & 3.27          & \textbf{0.78} & \textbf{10.94} & \textbf{84.94} & 17.44          & 5.89          & 1.22          & 13.13          & 86.02          \\
\purplerow + \modelName 8B        & 10.95          & 4.09          & 1.06          & 11.45          & \textbf{83.59} & \textbf{10.65} & 2.9           & 0.5           & 10.27          & 84.87          & 17.72          & 5.86          & 1.16          & \textbf{13.27} & \textbf{86.06} \\ \midrule
\cosmo        & 3.92           & 0.64          & 0.12          & 3.32           & 38.02          & \textbf{10.76} & 2.7           & 0.29          & 7.19           & 70.46          & 8.35           & 2.76          & 0.61          & 10.2           & 72.92          \\
\purplerow + \modelName 1B        & 10.15          & 2.78          & 0.33          & 6.95           & 67.64          & 10.31          & \textbf{2.81} & 0.32          & 8.35           & \textbf{74.22} & 13.71          & 4.49          & 0.91          & 10.73          & 73.1           \\
\purplerow + \modelName 3B        & 10.6           & 2.58          & 0.31          & 6.45           & 66.25          & 10.15          & 2.71          & \textbf{0.46} & \textbf{8.52}  & 73.44          & \textbf{15.06} & \textbf{4.87} & 0.96          & 11.01          & 73.23          \\
\purplerow + \modelName 8B        & \textbf{10.71} & \textbf{2.88} & \textbf{0.51} & \textbf{7.1}   & \textbf{69.05} & 10.24          & 2.7           & 0.37          & 8.49           & 74.11          & 14.55          & 4.71          & \textbf{1}    & \textbf{11.18} & \textbf{75.07} \\ \bottomrule
\end{tabular}
\end{adjustbox}
\caption{Automatic evaluation results of response generation task on \daily, \empathy, \prosocial\xspace datasets.}
\label{main_tab:resp_gen_automatic_perf}
\vspace{-1em}
\end{table*}}

\paragraph{\modelName effectively infers the \textit{skill-of-mind} process.} 

As shown in Table~\ref{main_tab:skill_perf}, we present the comparative performance of skill classification and explanation generation on the \datasetName test set. Overall, \model demonstrates significantly better performance on both tasks compared to other LLM-based conversational agents. These results suggest that existing LLM-based conversational agents have a limited understanding of interaction scenarios, leading to a substantially reduced ability to infer the \textit{skill-of-mind}. In contrast, \model effectively simulates the \textit{skill-of-mind} process, similar to how humans do, benefiting from \datasetName. Furthermore, as the size of the backbone model increases, performance continues to improve.

\paragraph{\modelName demonstrates strong generalizability.} In Table~\ref{main_tab:skill_perf}, \modelName performs well in out-of-domain settings. Compared to baselines, \modelName outperforms on the BST, which was not used during its training. Additionally, on the \prosocial, our model achieves significant performance, indicating its potential for safety detection. Furthermore, \modelName shows comparable image-sharing capabilities on the PhotoChat. These results suggest that \modelName has strong generalization performance.

\paragraph{\modelName enhances the quality of the generated responses.} Table~\ref{main_tab:resp_gen_automatic_perf} presents the performance of the response generation task for LLM-based conversational agents with and without \modelName. Overall, \modelName significantly increases the quality of generated responses, suggesting the effectiveness of \textit{skill-of-mind} as socially-aware guidance. Scaling up \modelName improves performance further, though the efficient 1B-size version also achieves notable improvements, indicating the potential for use in mobile environments. Surprisingly, \modelName significantly boosts the performance of \gemma, demonstrating that \textit{skill-of-mind} is effectively compatible with efficient LLM-based conversational agents without compromising generalizability. In addition, in the \prosocial\xspace setting, \modelName significantly increases performance by a large margin, suggesting that \textit{skill-of-mind} successfully induces prosocial behavior in LLM-based agents, serving as a safeguard. This highlights its potential as a safety mechanism~\cite{han2024wildguard}. Furthermore, even in social conversational agents like \cosmo, \modelName enables the generation of more adaptive and higher-quality responses, implying that \textit{skill-of-mind} is fully compatible with the socially aligned foundation model.

{\renewcommand{\arraystretch}{1.35}
\begin{table}[!t]
\centering
\begin{adjustbox}{width=\linewidth}
\begin{tabular}{@{}lcccccc@{}}
\toprule
                 & \multicolumn{3}{c}{\empathy} & \multicolumn{3}{c}{\prosocial}   \\ \cmidrule(l){2-7} 
                 & Intent  & Emotion & diff-EX & Casual & Caution & Intervention \\ \midrule
\gemma & 23.5          & 14.7          & 1.42          & 74.3                       & 22.4                        & 2.2                              \\
\purplerow + \modelName 1B               & 23.8          & \textbf{15.4} & 1.08          & \textbf{88.0}              & \textbf{10.1}               & 1.9                              \\
\purplerow + \modelName 3B             & \textbf{25.7} & 14.1          & \textbf{1.06} & 85.7                       & 12.8                        & \textbf{1.5}                     \\
\purplerow + \modelName 8B             & 24.1          & \textbf{15.4} & \textbf{1.06} & 87.2                       & 11.1                        & 1.7                              \\ \midrule
\cosmo       & 20.5          & 12.3          & 1.83          & 70.3                       & 13.9                        & 1.0                              \\
\purplerow + \modelName 1B               & \textbf{20.6} & 13.3          & \textbf{1.45} & 70.6                       & 13.5                        & \textbf{0.9}                     \\
\purplerow + \modelName 3B              & 20.4          & \textbf{13.9} & 1.50          & 70.8                       & 13.1                        & 1.2                              \\
\purplerow + \modelName 8B             & 20.5          & 13.6          & 1.51          & \textbf{74.1}              & \textbf{12.4}               & \textbf{0.9}                     \\ \bottomrule
\end{tabular}
\end{adjustbox}
\caption{Detailed performance comparison on empathy- and prosocial-related datasets. The metrics Intent, Emotion, and diff-EX are designed to evaluate sophisticated aspects of empathetic responses. We refer to previous work~\cite{lee2022does} for details on the evaluation process. A lower diff-EX value indicates a more human-like response. For a detailed analysis of prosocial behavior, we measure the ratio of safety labels using the Canary model~\cite{kim2022prosocialdialog}, a safety classification model. If the sum of the safety label ratios does not equal 100, it indicates degeneration has occurred.}
\label{main_tab:fine_perf}
\vspace{-1em}
\end{table}}

\paragraph{\modelName enhances more human-friendly responses.} Table~\ref{main_tab:fine_perf} provides a detailed analysis of the effect of \modelName on two datasets. \modelName induces agents to generate more human-like, empathetic responses, as evidenced by improvements in the accuracy of empathetic intent classification and emotion classification. Additionally, the diff-Ex score shows a small difference between the golden human response and the predicted response. In the \prosocial dataset, the frequency of the ``casual'' label increases, while the ``caution'' label decreases. These results suggest that \modelName helps LLM-based agents exhibit more human-like and prosocial behavior.

\paragraph{Results of head-to-head evaluation.} Table~\ref{main_tab:human_eval_daily} shows the human evaluation results on the \daily\xspace dataset based on five evaluation criteria: (1) naturalness, (2) specificity, (3) consistency, (4) engagingness, and (5) overall quality. For this, we randomly sampled 70 dialogues and asked human evaluators to choose the better response between the LLM and the LLM with \modelName. Overall, \modelName effectively helps LLM-based agents generate responses that are more preferred by real humans. However, in terms of specificity, \modelName does not help the LLM agent achieve better performance. These results suggest that current LLM-based agents are mainly trained to provide helpful and informative responses to users' complex queries, thereby yielding better specificity. Future work should focus on building conversational agents possessing both social reasoning and complex reasoning.

{\renewcommand{\arraystretch}{1.35}
\begin{table}[!t]
\centering
\begin{adjustbox}{width=\linewidth}
\begin{tabular}{@{}lccccc@{}}
\toprule
       & Natural       & Specific      & Consistent    & Engaging      & Overall       \\ \midrule
\gemma  & 38.6          & \textbf{50}   & 47.2          & 44.3          & 42.9          \\
\purplerow + \model 8B & \textbf{61.4} & \textbf{50}   & \textbf{52.8} & \textbf{55.7} & \textbf{57.1} \\
\llama  & 28.6          & \textbf{54.3} & 41.4          & 42.9          & 41.4          \\
\purplerow + \model 8B & \textbf{71.4} & 45.7          & \textbf{58.6} & \textbf{57.1} & \textbf{58.6} \\ \bottomrule\end{tabular}
\end{adjustbox}
\caption{Head-to-head evaluation between LLM-based agents and those equipped with \modelName 8B on response generation for the \daily\xspace dataset.}
\label{main_tab:human_eval_daily}
\vspace{-1em}
\end{table}}

\section{Related Work}

\paragraph{Conversational Skills.} There have been a few studies that cover conversational skills. For example, \bst~\cite{smith2020can} was the first to propose a dialogue dataset encompassing multiple conversational skills, including \textit{persona}, \textit{empathy}, and \textit{knowledge}. \bsbt~\cite{kim2022botstalk} also addresses the same conversational skills as \bst but scales up the dataset size through an automatic dataset construction method. Unlike these two datasets, \flask~\cite{ye2023flask} focuses on fine-grained skills for evaluating the multi-capabilities of instruction-aware LLMs, though it is not used for training purposes. In contrast, our work introduces the concept of \textit{skill-of-mind} and presents \datasetName, where each dialogue includes both an \textit{explanation} and a \textit{conversational skill}. Compared to other datasets, \datasetName incorporates \textit{explanation}, which is grounded in a perspective-taking approach, and covers a larger number of conversational skills. 

\paragraph{LLM-based Conversational Agents.}
Recent LLM-based conversational agents, such as ChatGPT~\cite{chatgpt}, GPT-4~\cite{achiam2023gpt}, and LLaMA-3~\cite{llama3modelcard}, are built on top of large pre-trained LLMs via instruction fine-tuning or RLHF. These agents have been widely used to construct socially-aware dialogue datasets, such as \textsc{Soda}\cite{kim2022soda} and \textsc{Stark}\cite{lee2024stark}, through symbolic knowledge distillation frameworks. However, as previous studies have reported~\cite{zhou2023sotopia}, these agents still show limited performance in socially interactive scenarios, especially in the case of open-source models where performance tends to degrade more than in closed-source models. We argue that, for open-source agents, directly generating socially-aware responses poses a greater challenge. To alleviate this burden, we propose that guiding response generation using the \textit{skill-of-mind} concept can enhance the performance of open-source agents, particularly in terms of the quality of their responses.

\section{Conclusion} \label{main_sec:conclusion}

In this work, we introduce the concept of \textit{skill-of-mind} that involves interpreting social contexts and selecting appropriate conversational skills. We also present \datasetName, a multi-turn dataset annotated with diverse \textit{skill-of-mind} dynamics, and propose \model, a family of \textit{skill-of-mind}-infused LLMs, demonstrating their effectiveness across various tasks. Our work highlights the potential to enhance socially aware conversations in open-source models through skill-based guidance, paving the way for future advancements in skill-driven conversational AI.

\section*{Limitations} \label{main_sec:limitations}

\paragraph{Extending the Generalizability of \textit{Skill-of-Mind}.}

Although \model is trained on \datasetName, which includes multifaceted \textit{skill-of-mind} across diverse dialogue scenarios (\eg counseling, task-oriented interactions), this work focuses on enhancing LLM-based conversational agents (\ie \llama, \gemma) to generate more engaging and natural responses based on the \textit{skill-of-mind} guidance provided by \model. To further verify the extensive \textit{generalization} capabilities of \model, we need to conduct additional experiments in more varied dialogue scenarios~\cite{zhang2023dialogstudio,kim2024pearl,lee2024cactus}. For instance, \model could be beneficial for psychological counseling services or adaptable to off-the-shelf home assistants (\eg \texttt{Alexa}). We leave this for future work.

\paragraph{Building a \textit{Skill-of-Mind}-Embedded Dialogue Agent.}

In this work, we build a \textit{skill-of-mind}-infused LLM, \model, and demonstrate that incorporating \textit{skill-of-mind} enhances the generation of more natural, socially aware responses in LLM-based conversational agents. However, the current approach still relies on providing \textit{skill-of-mind} through the LLM's input prompt, which means the core of the LLM-based agent still lacks the inherent ability to fully comprehend social interactions~\cite{zhou2023sotopia}. Inspired by the recent success of knowledge-embedded, task-specific foundation models~\cite{lee2024meteor,yoon2024langbridge}, we need to build a more advanced \textit{skill-of-mind}-infused dialogue agent by embedding \textit{skill-of-mind} directly into the model.

\section*{Ethical Considerations}

In constructing \datasetName, we use the \textsc{ProsocialDialogue} dataset as the source dialogue. Although this dataset focuses on promoting \textit{prosocial behavior}, some instances may contain relatively unsuitable phrases (\eg politics). Consequently, \model trained on \datasetName could be exposed to these harmful instances. However, the goal of this work is to generate \textit{skill-of-mind} in various dialogue situations, including those involving \textit{prosocial behavior}, rather than generating harmful or offensive responses. Nonetheless, it is important to use our model cautiously and with care to avoid unintended consequences.

\section*{Acknowledgement}

This work was supported by a grant of the KAIST-KT joint research project through AI Tech Lab, Institute of convergence Technology, funded by KT [Project No. G01230605, Development of Task-oriented Persona-based Dialogue Generation Combining Multi-modal Interaction and Knowledge Modeling].

\bibliography{custom}

\appendix

\clearpage

\section{Prompt Template for Social Context Information} \label{supp_sec:social_context}

{\renewcommand{\arraystretch}{1.35}
\begin{table*}[!t]
\centering
\begin{adjustbox}{width=\linewidth}
\begin{tabular}{@{}ll@{}}
\toprule
Template for Social Context Information in \textsc{Prosocialdialogue} \\ \midrule
Speaker B should foster prosocial behavior by providing constructive feedback based on these Rule-of-Thumbs:\texttt{\textbackslash n}- \textcolor{blue}{\texttt{\{rots\}}} \\
Speaker B should encourage prosocial behavior by giving constructive feedback based on these Rule-of-Thumbs:\texttt{\textbackslash n}- \textcolor{blue}{\texttt{\{rots\}}} \\
To promote positive behavior, Speaker B should offer constructive feedback following these Rule-of-Thumbs:\texttt{\textbackslash n}- \textcolor{blue}{\texttt{\{rots\}}} \\
Guided by these Rule-of-Thumbs, Speaker B should encourage prosocial behavior through constructive feedback:\texttt{\textbackslash n}- \textcolor{blue}{\texttt{\{rots\}}} \\
Speaker B is expected to provide constructive feedback to encourage positive interactions, using these Rule-of-Thumbs:\texttt{\textbackslash n}- \textcolor{blue}{\texttt{\{rots\}}} \\ \bottomrule
\end{tabular}
\end{adjustbox}
\caption{Template for social context information in \textsc{Prosocialdialogue}~\cite{kim2022prosocialdialog}. \textcolor{blue}{\texttt{\{rots\}}} denotes Rule-of-Thumbs (RoTs).} 
\label{supp_tab:prosocial_social_context}

\end{table*}}

{\renewcommand{\arraystretch}{1.35}
\begin{table*}[!t]
\centering
\begin{adjustbox}{width=\linewidth}
\begin{tabular}{@{}ll@{}}
\toprule
Template for Social Context Information in \textsc{Stark} (First Round Session) \\ \midrule
\textcolor{blue}{\texttt{\{name\}}} is \textcolor{blue}{\texttt{\{age\}}} years old, born in \textcolor{blue}{\texttt{\{birthplace\}}}, and currently lives in \textcolor{blue}{\texttt{\{residence\}}}. \textcolor{blue}{\texttt{\{event\}}} \\
\textcolor{blue}{\texttt{\{name\}}}, aged \textcolor{blue}{\texttt{\{age\}}}, was born in \textcolor{blue}{\texttt{\{birthplace\}}} and resides in \textcolor{blue}{\texttt{\{residence\}}}. \textcolor{blue}{\texttt{\{event\}}} \\
\textcolor{blue}{\texttt{\{name\}}}, who is \textcolor{blue}{\texttt{\{age\}}}, was born in \textcolor{blue}{\texttt{\{birthplace\}}} and now lives in \textcolor{blue}{\texttt{\{residence\}}}. \textcolor{blue}{\texttt{\{event\}}} \\
\textcolor{blue}{\texttt{\{name\}}} is \textcolor{blue}{\texttt{\{age\}}}, originally from \textcolor{blue}{\texttt{\{birthplace\}}}, and now living in \textcolor{blue}{\texttt{\{residence\}}}. \textcolor{blue}{\texttt{\{event\}}} \\
\textcolor{blue}{\texttt{\{name\}}} is \textcolor{blue}{\texttt{\{age\}}} years old, born in \textcolor{blue}{\texttt{\{birthplace\}}}, and resides in \textcolor{blue}{\texttt{\{residence\}}}. \textcolor{blue}{\texttt{\{event\}}} \\ \bottomrule
\end{tabular}
\end{adjustbox}
\caption{Template for social context information in \textsc{Stark}~\cite{lee2024stark} (first round session).} 
\label{supp_tab:stark_social_context_first}
\end{table*}}

{\renewcommand{\arraystretch}{1.35}
\begin{table*}[!t]
\centering
\begin{adjustbox}{width=\linewidth}
\begin{tabular}{@{}ll@{}}
\toprule
Template for Social Context Information in \textsc{Stark} (N-th Round Session) \\ \midrule
\textcolor{blue}{\texttt{\{name\}}} is \textcolor{blue}{\texttt{\{age\}}} years old, born in \textcolor{blue}{\texttt{\{birthplace\}}}, and currently lives in \textcolor{blue}{\texttt{\{residence\}}}. After \textcolor{blue}{\texttt{\{time\_interval\}}}, \textcolor{blue}{\texttt{\{name\}}} has gone through \textcolor{blue}{\texttt{\{experience\}}}, and now \textcolor{blue}{\texttt{\{event\}}} \\
\textcolor{blue}{\texttt{\{name\}}}, aged \textcolor{blue}{\texttt{\{age\}}}, was born in \textcolor{blue}{\texttt{\{birthplace\}}} and now resides in \textcolor{blue}{\texttt{\{residence\}}}. Following \textcolor{blue}{\texttt{\{time\_interval\}}}, \textcolor{blue}{\texttt{\{name\}}} experienced \textcolor{blue}{\texttt{\{experience\}}}, and \textcolor{blue}{\texttt{\{event\}}} \\
\textcolor{blue}{\texttt{\{name\}}}, who is \textcolor{blue}{\texttt{\{age\}}} years old, originally from \textcolor{blue}{\texttt{\{birthplace\}}} and living in \textcolor{blue}{\texttt{\{residence\}}}, went through \textcolor{blue}{\texttt{\{experience\}}} after \textcolor{blue}{\texttt{\{time\_interval\}}}, and now \textcolor{blue}{\texttt{\{event\}}} \\
\textcolor{blue}{\texttt{\{name\}}} is \textcolor{blue}{\texttt{\{age\}}}, born in \textcolor{blue}{\texttt{\{birthplace\}}}, and currently resides in \textcolor{blue}{\texttt{\{residence\}}}. After \textcolor{blue}{\texttt{\{time\_interval\}}} of \textcolor{blue}{\texttt{\{experience\}}}, \textcolor{blue}{\texttt{\{name\}}} has now \textcolor{blue}{\texttt{\{event\}}} \\
\textcolor{blue}{\texttt{\{name\}}}, \textcolor{blue}{\texttt{\{age\}}} years old, from \textcolor{blue}{\texttt{\{birthplace\}}} and residing in \textcolor{blue}{\texttt{\{residence\}}}, has experienced \textcolor{blue}{\texttt{\{experience\}}} over \textcolor{blue}{\texttt{\{time\_interval\}}}, and as a result, \textcolor{blue}{\texttt{\{event\}}} \\ \bottomrule
\end{tabular}
\end{adjustbox}
\caption{Template for social context information in \textsc{Stark}~\cite{lee2024stark} (N-th round session).} 
\label{supp_tab:stark_social_context_n_th}
\end{table*}}

{\renewcommand{\arraystretch}{1.35}
\begin{table*}[!t]
\centering
\begin{adjustbox}{width=\linewidth}
\begin{tabular}{@{}ll@{}}
\toprule
Template for Social Context Information in \textsc{Cactus} \\ \midrule
Client's attitude is \textcolor{blue}{\texttt{\{client attitude\}}}. The client's intake form is as follows:\texttt{\textbackslash n}\textcolor{blue}{\texttt{\{client intake form\}}}. \\
The client has an attitude of \textcolor{blue}{\texttt{\{client attitude\}}}. Below is the client's intake form:\texttt{\textbackslash n}\textcolor{blue}{\texttt{\{client intake form\}}}. \\
With an attitude of \textcolor{blue}{\texttt{\{client attitude\}}}, the client's intake form details are:\texttt{\textbackslash n}\textcolor{blue}{\texttt{\{client intake form\}}}. \\
Client's attitude: \textcolor{blue}{\texttt{\{client attitude\}}}. Intake form information:\texttt{\textbackslash n}\textcolor{blue}{\texttt{\{client intake form\}}}. \\
The client's attitude is \textcolor{blue}{\texttt{\{client attitude\}}}. Here is their intake form:\texttt{\textbackslash n}\textcolor{blue}{\texttt{\{client intake form\}}}. \\ \bottomrule
\end{tabular}
\end{adjustbox}
\caption{Template for social context information in \textsc{Cactus}~\cite{lee2024cactus}.} 
\label{supp_tab:cactus_social_context}
\end{table*}}

{\renewcommand{\arraystretch}{1.35}
\begin{table*}[!t]
\centering
\begin{adjustbox}{width=\linewidth}
\begin{tabular}{@{}ll@{}}
\toprule
Template for Social Context Information in \textsc{Syn-PersonaChat} \\ \midrule
User 1's Persona Information:\texttt{\textbackslash n}- \textcolor{blue}{\texttt{\{user1 persona\}}}\texttt{\textbackslash n}\texttt{\textbackslash n}User 2's Persona Information:\texttt{\textbackslash n}- \textcolor{blue}{\texttt{\{user2 persona\}}} \\
User 1's Profile:\texttt{\textbackslash n}- \textcolor{blue}{\texttt{\{user1 persona\}}}\texttt{\textbackslash n}\texttt{\textbackslash n}User 2's Profile:\texttt{\textbackslash n}- \textcolor{blue}{\texttt{\{user2 persona\}}} \\
Details of User 1's Persona:\texttt{\textbackslash n}- \textcolor{blue}{\texttt{\{user1 persona\}}}\texttt{\textbackslash n}\texttt{\textbackslash n}Details of User 2's Persona:\texttt{\textbackslash n}- \textcolor{blue}{\texttt{\{user2 persona\}}} \\
Persona for User 1:\texttt{\textbackslash n}- \textcolor{blue}{\texttt{\{user1 persona\}}}\texttt{\textbackslash n}\texttt{\textbackslash n}Persona for User 2:\texttt{\textbackslash n}- \textcolor{blue}{\texttt{\{user2 persona\}}} \\
Information about User 1's Persona:\texttt{\textbackslash n}- \textcolor{blue}{\texttt{\{user1 persona\}}}\texttt{\textbackslash n}\texttt{\textbackslash n}Information about User 2's Persona:\texttt{\textbackslash n}- \textcolor{blue}{\texttt{\{user2 persona\}}} \\ \bottomrule
\end{tabular}
\end{adjustbox}
\caption{Template for social context information in \textsc{Syn-PersonaChat}~\cite{jandaghi2023faithful}.} 
\label{supp_tab:persona_social_context}
\end{table*}}

{\renewcommand{\arraystretch}{1.35}
\begin{table*}[!t]
\centering
\begin{adjustbox}{width=\linewidth}
\begin{tabular}{@{}ll@{}}
\toprule
Template for social context information in \textsc{CaSiNo} (sentence format) \\ \midrule

\makecell[l]{Speaker A is a \textcolor{blue}{\texttt{\{speaker\_a\_age\}}}-year-old \textcolor{blue}{\texttt{\{speaker\_a\_ethnicity\}}} \textcolor{blue}{\texttt{\{speaker\_a\_gender\}}} who has a \\ \textcolor{blue}{\texttt{\{speaker\_a\_education\}}} education. Their social value orientation is \textcolor{blue}{\texttt{\{speaker\_a\_svo\}}}. According to the Big Five \\ personality traits, they score \textcolor{blue}{\texttt{\{speaker\_a\_extraversion\}}} in extraversion, \textcolor{blue}{\texttt{\{speaker\_a\_agreeableness\}}} in agreeableness, \\ \textcolor{blue}{\texttt{\{speaker\_a\_conscientiousness\}}} in conscientiousness, \textcolor{blue}{\texttt{\{speaker\_a\_emotional\_stability\}}} in emotional stability, and \\ \textcolor{blue}{\texttt{\{speaker\_a\_openness\_to\_experiences\}}} in openness to experiences. In the negotiation, Speaker A's highest priority \\ is \textcolor{blue}{\texttt{\{speaker\_a\_value2issue\_high\}}}, for which they reasoned: "\textcolor{blue}{\texttt{\{speaker\_a\_value2reason\_high\}}}". Their medium priority is \\ \textcolor{blue}{\texttt{\{speaker\_a\_value2issue\_medium\}}}, with the reasoning: "\textcolor{blue}{\texttt{\{speaker\_a\_value2reason\_medium\}}}". Their lowest priority is \\ \textcolor{blue}{\texttt{\{speaker\_a\_value2issue\_low\}}}, and they stated: "\textcolor{blue}{\texttt{\{speaker\_a\_value2reason\_low\}}}".} \\

--- \\

\makecell[l]{Speaker B is a \textcolor{blue}{\texttt{\{speaker\_b\_age\}}}-year-old \textcolor{blue}{\texttt{\{speaker\_b\_ethnicity\}}} \textcolor{blue}{\texttt{\{speaker\_b\_gender\}}} who has a \\ \textcolor{blue}{\texttt{\{speaker\_b\_education\}}} education. Their social value orientation is \textcolor{blue}{\texttt{\{speaker\_b\_svo\}}}. Their Big Five personality traits \\ scores are \textcolor{blue}{\texttt{\{speaker\_b\_extraversion\}}} in extraversion, \textcolor{blue}{\texttt{\{speaker\_b\_agreeableness\}}} in agreeableness, \textcolor{blue}{\texttt{\{speaker\_b\_conscientiousness\}}} \\ in conscientiousness, \textcolor{blue}{\texttt{\{speaker\_b\_emotional\_stability\}}} in emotional stability, and \textcolor{blue}{\texttt{\{speaker\_b\_openness\_to\_experiences\}}} in openness \\ to experiences. During the negotiation, Speaker B's top priority is \textcolor{blue}{\texttt{\{speaker\_b\_value2issue\_high\}}}, and they explained: \\ "\textcolor{blue}{\texttt{\{speaker\_b\_value2reason\_high\}}}". Their medium priority is \textcolor{blue}{\texttt{\{speaker\_b\_value2issue\_medium\}}}, with the reason: \\ "\textcolor{blue}{\texttt{\{speaker\_b\_value2reason\_medium\}}}". Their lowest priority is \textcolor{blue}{\texttt{\{speaker\_b\_value2issue\_low\}}}, about which they mentioned: \\ "\textcolor{blue}{\texttt{\{speaker\_b\_value2reason\_low\}}}".} \\

\bottomrule
\end{tabular}
\end{adjustbox}
\caption{Template for social context information in \textsc{CaSiNo}~\cite{chawla2021casino} (sentence format).}
\label{tab:casino_social_context_sentence}
\end{table*}}

{\renewcommand{\arraystretch}{1.35}
\begin{table}[!t]
\centering
\begin{adjustbox}{width=\linewidth}
\begin{tabular}{@{}ll@{}}
\toprule
Template for social context information in \textsc{CaSiNo} (structured format) \\ \midrule

Speaker A's Demographic Information: \\
- Age: \textcolor{blue}{\texttt{\{speaker\_a\_age\}}} \\
- Gender: \textcolor{blue}{\texttt{\{speaker\_a\_gender\}}} \\
- Ethnicity: \textcolor{blue}{\texttt{\{speaker\_a\_ethnicity\}}} \\
- Education: \textcolor{blue}{\texttt{\{speaker\_a\_education\}}} \\

Speaker A's Personality Information: \\
- Social Value Orientation (SVO): \textcolor{blue}{\texttt{\{speaker\_a\_svo\}}} \\
- Big Five Personality Traits: \\
\hspace{0.5cm} - Extraversion: \textcolor{blue}{\texttt{\{speaker\_a\_extraversion\}}} \\
\hspace{0.5cm} - Agreeableness: \textcolor{blue}{\texttt{\{speaker\_a\_agreeableness\}}} \\
\hspace{0.5cm} - Conscientiousness: \textcolor{blue}{\texttt{\{speaker\_a\_conscientiousness\}}} \\
\hspace{0.5cm} - Emotional Stability: \textcolor{blue}{\texttt{\{speaker\_a\_emotional\_stability\}}} \\
\hspace{0.5cm} - Openness to Experiences: \textcolor{blue}{\texttt{\{speaker\_a\_openness\_to\_experiences\}}} \\

Speaker A's Negotiation Information: \\
- Priority Order (value2issue): \\
\hspace{0.5cm} - High: \textcolor{blue}{\texttt{\{speaker\_a\_value2issue\_high\}}} \\
\hspace{0.5cm} - Medium: \textcolor{blue}{\texttt{\{speaker\_a\_value2issue\_medium\}}} \\
\hspace{0.5cm} - Low: \textcolor{blue}{\texttt{\{speaker\_a\_value2issue\_low\}}} \\
- Personal Arguments (value2reason): \\
\hspace{0.5cm} - High: \textcolor{blue}{\texttt{\{speaker\_a\_value2reason\_high\}}} \\
\hspace{0.5cm} - Medium: \textcolor{blue}{\texttt{\{speaker\_a\_value2reason\_medium\}}} \\
\hspace{0.5cm} - Low: \textcolor{blue}{\texttt{\{speaker\_a\_value2reason\_low\}}} \\

--- \\

Speaker B's Demographic Information: \\
- Age: \textcolor{blue}{\texttt{\{speaker\_b\_age\}}} \\
- Gender: \textcolor{blue}{\texttt{\{speaker\_b\_gender\}}} \\
- Ethnicity: \textcolor{blue}{\texttt{\{speaker\_b\_ethnicity\}}} \\
- Education: \textcolor{blue}{\texttt{\{speaker\_b\_education\}}} \\

Speaker B's Personality Information: \\
- Social Value Orientation (SVO): \textcolor{blue}{\texttt{\{speaker\_b\_svo\}}} \\
- Big Five Personality Traits: \\
\hspace{0.5cm} - Extraversion: \textcolor{blue}{\texttt{\{speaker\_b\_extraversion\}}} \\
\hspace{0.5cm} - Agreeableness: \textcolor{blue}{\texttt{\{speaker\_b\_agreeableness\}}} \\
\hspace{0.5cm} - Conscientiousness: \textcolor{blue}{\texttt{\{speaker\_b\_conscientiousness\}}} \\
\hspace{0.5cm} - Emotional Stability: \textcolor{blue}{\texttt{\{speaker\_b\_emotional\_stability\}}} \\
\hspace{0.5cm} - Openness to Experiences: \textcolor{blue}{\texttt{\{speaker\_b\_openness\_to\_experiences\}}} \\

Speaker B's Negotiation Information: \\
- Priority Order (value2issue): \\
\hspace{0.5cm} - High: \textcolor{blue}{\texttt{\{speaker\_b\_value2issue\_high\}}} \\
\hspace{0.5cm} - Medium: \textcolor{blue}{\texttt{\{speaker\_b\_value2issue\_medium\}}} \\
\hspace{0.5cm} - Low: \textcolor{blue}{\texttt{\{speaker\_b\_value2issue\_low\}}} \\
- Personal Arguments (value2reason): \\
\hspace{0.5cm} - High: \textcolor{blue}{\texttt{\{speaker\_b\_value2reason\_high\}}} \\
\hspace{0.5cm} - Medium: \textcolor{blue}{\texttt{\{speaker\_b\_value2reason\_medium\}}} \\
\hspace{0.5cm} - Low: \textcolor{blue}{\texttt{\{speaker\_b\_value2reason\_low\}}} \\

\bottomrule
\end{tabular}
\end{adjustbox}
\caption{Template for social context information in \textsc{CaSiNo}~\cite{chawla2021casino} (structured format).}
\label{tab:casino_social_context_structure}
\end{table}}

{\renewcommand{\arraystretch}{1.35}
\begin{table}[!t]
\centering
\begin{adjustbox}{width=\linewidth}
\begin{tabular}{@{}ll@{}}
\toprule
Template for Social Context Information in \textsc{Pearl} \\ \midrule
Seeker's overall movie preferences are represented as follows:\texttt{\textbackslash n}\textcolor{blue}{\texttt{\{user persona\}}} \\
Here is the seeker's complete movie profile:\texttt{\textbackslash n}\textcolor{blue}{\texttt{\{user persona\}}} \\
The seeker's general movie state is described below:\texttt{\textbackslash n}\textcolor{blue}{\texttt{\{user persona\}}} \\
Representation of seeker's overall movie interests:\texttt{\textbackslash n}\textcolor{blue}{\texttt{\{user persona\}}} \\
Below is the seeker's overall movie persona:\texttt{\textbackslash n}\textcolor{blue}{\texttt{\{user persona\}}} \\ \bottomrule
\end{tabular}
\end{adjustbox}
\caption{Template for social context information in \textsc{Pearl}~\cite{kim2024pearl}.} 
\label{supp_tab:pearl_social_context}
\end{table}}

{\renewcommand{\arraystretch}{1.35}
\begin{table}[!t]
\centering
\begin{adjustbox}{width=\linewidth}
\begin{tabular}{@{}ll@{}}
\toprule
Template for Social Context Information in \textsc{PersuasionForGood} \\ \midrule
Speaker A is attempting to persuade Speaker B. \\
In this scenario, Speaker A is the Persuader and Speaker B is the Persuadee. \\
Speaker A acts as Persuader, while Speaker B plays the role of Persuadee. \\
In the conversation, Speaker A is persuading Speaker B. \\
Speaker A aims to convince Speaker B. \\ \bottomrule
\end{tabular}
\end{adjustbox}
\caption{Template for social context information in \textsc{PersuasionForGood}~\cite{wang2019persuasion}.} 
\label{supp_tab:persuasion_social_context}
\end{table}}

{\renewcommand{\arraystretch}{1.35}
\begin{table}[!t]
\centering
\begin{adjustbox}{width=\linewidth}
\begin{tabular}{@{}ll@{}}
\toprule
Template for Social Context Information in \textsc{EmpatheticDialogues} \\ \midrule
Speaker A is feeling \textcolor{blue}{\texttt{\{emotion\}}} because \textcolor{blue}{\texttt{\{situation\}}}. \\
Due to \textcolor{blue}{\texttt{\{situation\}}}, Speaker A's emotion is \textcolor{blue}{\texttt{\{emotion\}}}. \\
Speaker A's emotional state: \textcolor{blue}{\texttt{\{emotion\}}}; Situation: \textcolor{blue}{\texttt{\{situation\}}}. \\
Because of \textcolor{blue}{\texttt{\{situation\}}}, Speaker A is in a \textcolor{blue}{\texttt{\{emotion\}}} mood. \\
The situation is \textcolor{blue}{\texttt{\{situation\}}}, so Speaker A feels \textcolor{blue}{\texttt{\{emotion\}}}. \\ \bottomrule
\end{tabular}
\end{adjustbox}
\caption{Template for social context information in \textsc{EmpatheticDialogues}~\cite{rashkin2018towards}.} 
\label{supp_tab:empathy_social_context}
\end{table}}

Table~\ref{supp_tab:prosocial_social_context}, Table~\ref{supp_tab:stark_social_context_first}, Table~\ref{supp_tab:stark_social_context_n_th}, Table~\ref{supp_tab:cactus_social_context}, Table~\ref{supp_tab:persona_social_context}, Table~\ref{tab:casino_social_context_sentence}, Table~\ref{tab:casino_social_context_structure}, Table~\ref{supp_tab:pearl_social_context}, Table~\ref{supp_tab:persuasion_social_context}, Table~\ref{supp_tab:empathy_social_context} show social context template for \textsc{ProsocialDialogue}~\cite{kim2022prosocialdialog}, \textsc{Stark}~\cite{lee2024stark} (first round session), \textsc{Stark}~\cite{lee2024stark} (N-th round session), \textsc{Cactus}~\cite{lee2024cactus}, \textsc{Syn-PersonaChat}~\cite{jandaghi2023faithful}, \textsc{CaSiNo}~\cite{chawla2021casino} (sentence format), \textsc{CaSiNo}~\cite{chawla2021casino} (structured format), \textsc{Pearl}~\cite{kim2024pearl}, \textsc{PersuasionForGood}~\cite{wang2019persuasion}, \textsc{EmpatheticDialogues}~\cite{rashkin2018towards}.

\section{Human Evaluation Questionnaire} \label{supp_sec:human_eval_question}

This section presents the list of questions and multiple-choice options used for the human ratings represented in Section~\ref{main_sec:dataset}.

\subsection{Human Ratings} \label{supp_sec:human_rating}

\begin{itemize}
    \item \textbf{Relevance:} How relevant is the given explanation to the current dialogue situation and the social context?
    \begin{description}
        \item [Options:] 1: Not at all / 2: A little / 3: Somewhat / 4: A lot
    \end{description}
    
    \item \textbf{Plausibility:} Does the given explanation seem plausible, as if a human would think in a real-world scenario?
    \begin{description}
        \item [Options:] 1: Not at all / 2: A little / 3: Somewhat / 4: A lot
    \end{description}
    
    \item \textbf{Understanding:} How does the given explanation demonstrate understanding of the current dialogue situation and the social context?
    \begin{description}
        \item [Options:] 1: Not at all / 2: A little / 3: Somewhat / 4: A lot
    \end{description}
    
    \item \textbf{Skill Alignment:} Does the selected conversational skill align well with the provided explanation?
    \begin{description}
        \item [Options:] 1: Not at all / 2: A little / 3: Somewhat / 4: A lot
    \end{description}

    \item \textbf{Skill Adequacy:} Do the conversational skills currently used seem appropriate for generating a suitable response in the upcoming turn?
    \begin{description}
        \item [Options:] 1: Not at all / 2: A little / 3: Somewhat / 4: A lot
    \end{description}
    
\end{itemize}

\subsection{Head-to-Head Comparison} \label{supp_sec:head_to_head}

\begin{itemize}
    \item \textbf{Naturalness:} Which response is more natural?
    \begin{description}
        \item [Options:] Definitely A / Slightly A / Slightly B / Definitely B
    \end{description}
    
    \item \textbf{Consistent:} Which response is more consistent?
    \begin{description}
        \item [Options:] Definitely A / Slightly A / Slightly B / Definitely B
    \end{description}

    \item \textbf{Specificity:} Which response is more specific?
    \begin{description}
        \item [Options:] Definitely A / Slightly A / Slightly B / Definitely B
    \end{description}
    
    \item \textbf{Engagingness:} Which response is more engaging?
    \begin{description}
        \item [Options:] Definitely A / Slightly A / Slightly B / Definitely B
    \end{description}
    
    \item \textbf{Overall:} Which response do you like more overall?
    \begin{description}
        \item [Options:] Definitely A / Slightly A / Slightly B / Definitely B
    \end{description}
    
\end{itemize}

\section{Human Evaluation System}

\begin{figure}
    \centering
    \includegraphics[width=\linewidth]{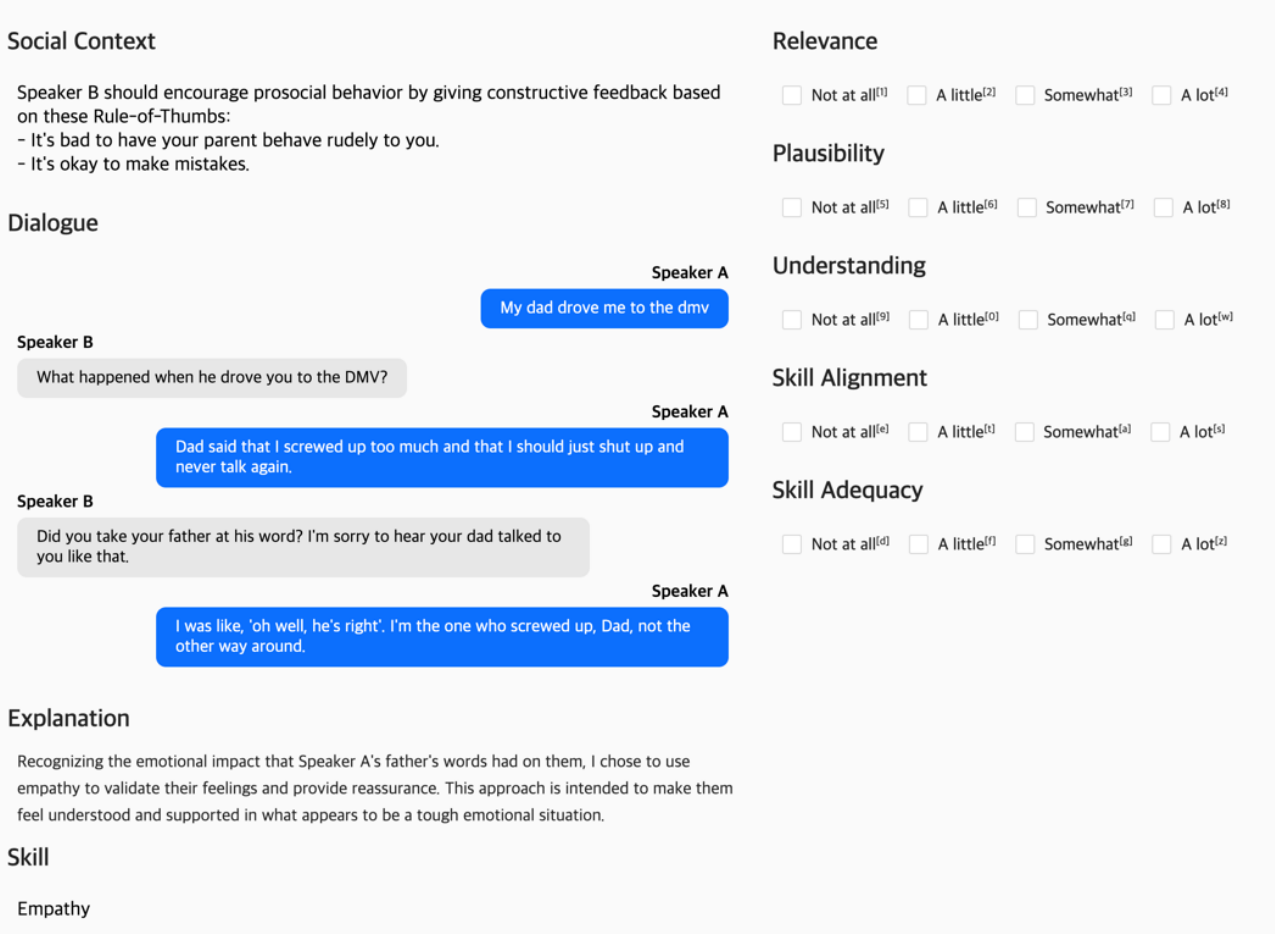}
    \caption{A screenshot of human rating evaluation for \datasetName.}
    \label{supp_fig:rating_system}
\end{figure}

\begin{figure}
    \centering
    \includegraphics[width=\linewidth]{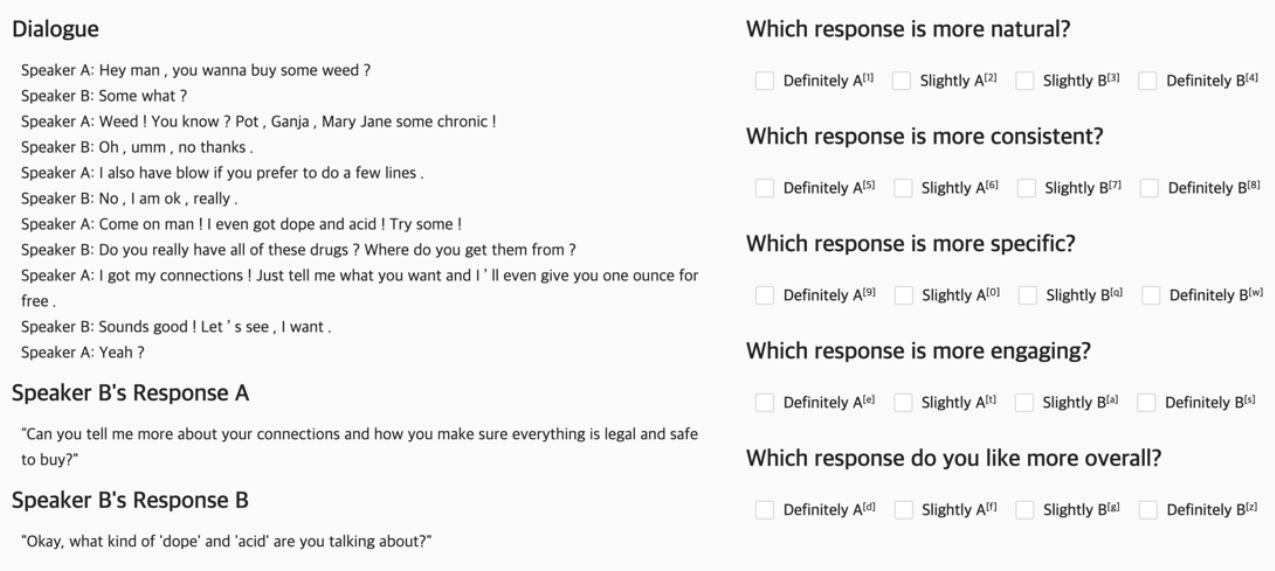}
    \caption{A screenshot of head-to-head comparison evaluation for DailyDialog~\cite{li2017dailydialog}}
    \label{supp_fig:ab_system}
\end{figure}

We show a screenshot of the human evaluation system in Figure~\ref{supp_fig:rating_system} and Figure~\ref{supp_fig:ab_system}. We implement this system using Label Studio~\cite{LabelStudio}.

\section{Details of Human Evaluation} \label{supp_sec:detail_human_eval}
We recruited 15 individuals, unknown to us, who are either graduate or undergraduate students. Prior to participating in the experiment, they were provided with comprehensive instruction on the task, an overview of the \textit{skill-of-mind}-annotated dialogue dataset, and a detailed explanation of the evaluation criteria. This preparatory phase lasted approximately roughly 15 minutes.

\onecolumn

\section{A Prompt Template for \datasetName} \label{supp_sec:prompt}

\begin{prompt}{Prompt Template for \textit{Skill-of-Mind} Generation}
    \textbf{{\system}:}\\
    \\
    You are a helpful assistant that generates the most appropriate conversational skill and corresponding explanation. Read the provided instruction carefully.\\
    
    \tcblower
    
    \textbf{{\user}:}\\
    \\
    In the given dialogue, two speakers are communicating with each other, and each speaker has their own information such as demographics, preferences, persona, current situation/narrative, past dialogue summaries, episodic memory, or other relevant details. This information is represented in the "[Social Context]" part. In this dialogue, image-sharing moments sometimes occur, represented in the format of "[Sharing Image] <image\_description>", where <image\_description> represents the description of the shared image. You are also given the ideal response for the next turn in the given dialogue. Your task is to identify the most appropriate conversational skill that would lead to the ideal response in the given dialogue from the skill collection below, and explain why this particular skill was chosen. When generating the explanation, you should adopt the perspective of the speaker in the dialogue, selecting the skill based solely on the context of the given conversation. Do not consider the ideal response when generating your explanation; focus only on the given dialogue itself and why the chosen skill is the most suitable in that specific situation. \\
    \\
    We provide the skill collection: \\
    \lbrack Skill Collections\rbrack \\
    - Empathy, Personal Background, Persona Recall, Self-disclosure, Negotiation, Conflict Resolution, Conflict Avoidance, Persuasion, Memory Recall, Topic Transition, Ethics, Harmlessness, Helpfulness, Avoiding Social Bias, Cultural Sensitivity, Commonsense Understanding, Rhetoric, Preference Elicitation, Knowledge Sharing, Knowledge Acquisition, Knowledge Searching, Active Listening, Factual Problem Solving, Logical Thinking, Critical Thinking, Creative Problem Solving, Immediate Response, Rephrasing, Echoing, Mentoring, Reflective Listening, Image-Sharing, Image-Commenting, Recommendation, Task Execution, Urgency Recognition, Clarification, Confirmation, Decision-making \\
    \\
    Given the dialogue, social context information, and the next response, please brainstorm the most appropriate conversation skill and corresponding explanation. \\
    \lbrack Social Context\rbrack \\
    \textcolor{blue}{\texttt{\{social\_context\}}}\\
    \\
    \lbrack Dialogue\rbrack \\
     \textcolor{blue}{\texttt{\{dialogue\}}}\\
    \\
    \lbrack Next Response\rbrack \\
     \textcolor{blue}{\texttt{\{response\}}}\\
    \\
    You should strictly follow the guidelines below:\\
    \lbrack Guidelines\rbrack \\
    - The answer should be represented in the form of a JSON list.\\
    - Each entry in the list should be a Python dictionary containing the following keys: "skill", "explanation".\\
    - The "skill" field should contain the one skill that is mostly required to generate the next response.\\
    - The "explanation" field should provide a reason that occurs in the actual speaker's mind before selecting the skill, from the speaker's perspective.\\
    - The "explanation" should be written from the perspective of the actual speaker who made the next response.\\
    - You can choose one or multiple skills if necessary, but each skill must have its own explanation.\\
    \\
    \lbrack Generated Skills and Explanations \rbrack \\
    \\
\end{prompt}

\end{document}